\documentclass[letterpaper, 10 pt, conference]{ieeeconf}  

\IEEEoverridecommandlockouts                              

\usepackage{graphics} 
\usepackage{epsfig} 
\usepackage{mathptmx} 
\usepackage{times} 
\usepackage{amsmath} 
\usepackage{amssymb}  
\usepackage{multirow}
\usepackage{graphicx}
\usepackage{caption}
\usepackage{cite}
\usepackage{hyperref} 
\usepackage{booktabs} 

\usepackage{xcolor}

\title{\LARGE \bf
Know your limits! Optimize the robot's behavior through self-awareness}

\author{Esteve Valls Mascar\'o$^{1}$ and Dongheui Lee$^{1,2}$
\thanks{$^{1}$Esteve Valls Mascaro and Dongheui Lee are with Autonomous Systems, Technische Universität Wien (TU Wien), Vienna, Austria (e-mail: \texttt{\{esteve.valls.mascaro, dongheui.lee\}@tuwien.ac.at}).}%
\thanks{$^{2}$Dongheui Lee is also with the Institute of Robotics and Mechatronics (DLR), German Aerospace Center, Wessling, Germany.}\\%
\href{https://evm7.github.io/Self-AWare/}{\color{blue} evm7.github.io/Self-AWare}
}

\begin{document}

\thispagestyle{empty}
\pagestyle{empty}

\twocolumn[{%
\renewcommand\twocolumn[1][]{#1}%
\maketitle
\begin{center}
    \centering
    \vspace{-6mm}
    \captionsetup{type=figure}
    \includegraphics[width=.98\textwidth]{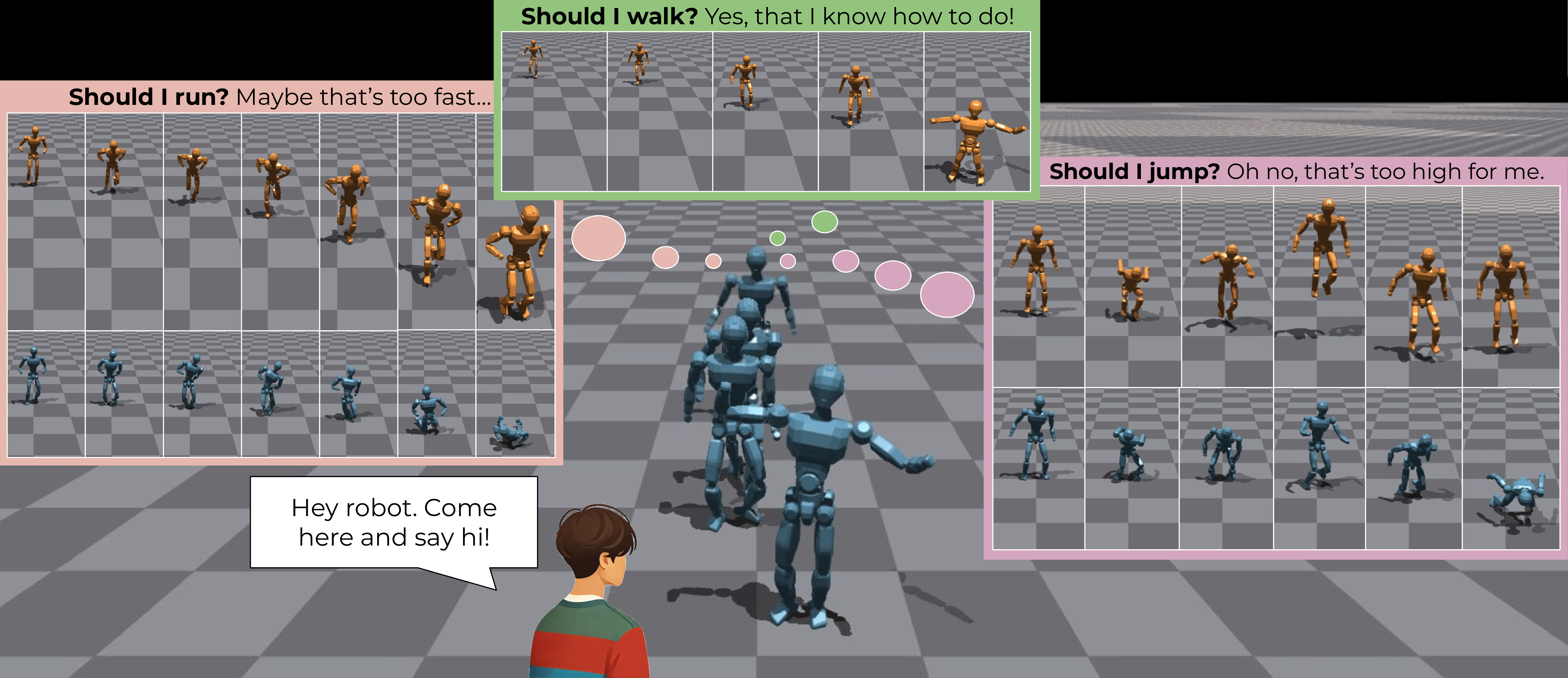}
    \captionof{figure}{\textbf{Optimizing robot behavior control from high-level task commands using self-awareness}. Given an instruction by an operator, the robot generates multiple potential behaviors to accomplish the task, evaluates them based on its capabilities and limits, and selects the most suitable one to execute. This work introduces a motion adaptation with a Self-AWare model (SAW) to anticipate how well a robot can follow a given reference by ranking multiple potential behaviors and choosing the optimal one. For instance, in a scenario with three potential actions—walking, running, and jumping—the robot assesses each option and determines that walking is the most appropriate. Consequently, the robot walks to the person and says, "Hi." In this image, the generated references are shown by orange robots, while the robot's behavior when attempting to follow them is depicted in blue. Note that the reference motions are fixed just for visualizations, and do not consider gravity.}
    \label{fig:overview}
\end{center}%
}]
\renewcommand{\thefootnote}{\fnsymbol{footnote}}
\footnotetext{$^{1}$Esteve Valls Mascaro and Dongheui Lee are with Autonomous Systems, Technische Universität Wien (TU Wien), Vienna, Austria (e-mail: \texttt{\{esteve.valls.mascaro, dongheui.lee\}@tuwien.ac.at}).}

\footnotetext{$^{2}$Dongheui Lee is also with the Institute of Robotics and Mechatronics (DLR), German Aerospace Center, Wessling, Germany.}

\begin{abstract}

As humanoid robots transition from labs to real-world environments, it is essential to democratize robot control for non-expert users. Recent human-robot imitation algorithms focus on following a reference human motion with high precision, but they are susceptible to the quality of the reference motion and require the human operator to simplify its movements to match the robot's capabilities. Instead, we consider that the robot should understand and adapt the reference motion to its own abilities, facilitating the operator's task. For that, we introduce a deep-learning model that anticipates the robot's performance when imitating a given reference. Then, our system can generate multiple references given a high-level task command, assign a score to each of them, and select the best reference to achieve the desired robot behavior. Our Self-AWare model (SAW) ranks potential robot behaviors based on various criteria, such as fall likelihood, adherence to the reference motion, and smoothness. We integrate advanced motion generation, robot control, and SAW in one unique system, ensuring optimal robot behavior for any task command. For instance, SAW can anticipate falls with 99.29\% accuracy.

\end{abstract}

\section{INTRODUCTION}
Imagine a world where bipedal robots walk among us, mimicking human movements and behaviors with remarkable precision. Such robots could revolutionize industries ranging from healthcare to disaster response, offering unparalleled assistance and efficiency.  However, teaching robots how to robustly imitate human behaviors is very challenging, especially for bipedal robots. Recent works \cite{cheng2024express, he2024learning, rl_bipedal1, rl_bipedal2, zhang2024whole, zhang2024wococo, 2024ictrl} have shown promising potential by training a robot to imitate a given human reference using goal-conditioned reinforcement learning (RL) \cite{goalcond_rl}. However, exactly following a human reference motion might fall out of the robot's capabilities. To address this issue, previous works filtered out unfeasible motions and limited the commanded behaviors to upper body movements \cite{cheng2024express, he2024learning} or one specific task \cite{rl_bipedal1, rl_bipedal2, zhang2024whole, zhang2024wococo}.  However, this strategy presents two main problems: (a) Should the human operator adapt its reference behavior to the robot's abilities and constraints? (b) How can robots handle dynamic motions that are out of their expertise? This paper is motivated to answer those questions by doting the robots with self-awareness: the ability to understand any motion command and adapt it according to their own limitations and capabilities.

As humans, we inherently possess self-awareness, which helps us recognize our physical and cognitive boundaries. For instance, a novice parkour enthusiast understands that attempting a high-risk maneuver without sufficient skill can lead to injury. This self-awareness guides them to modify their actions, opting for safer, more manageable moves until they gain more experience and confidence. Similarly, self-awareness in robots is equally important to ensure safety, efficiency, and high task performance. Without it, robots might blindly try to imitate human motions without considering their own physical constraints, leading to falls, collisions, or mechanical damage. In fact, self-awareness should also entail the robot's expertise: a robot might be able to jump with one leg, but still has not learned how. Therefore, during execution, the robot should ideally explore how to move like a human jumping with one leg and adapt its movement based on its own capabilities. By incorporating self-awareness, robots can assess their capabilities and make informed decisions, preventing actions beyond their abilities, as shown in Fig. \ref{fig:overview}. In this paper, we reformulate this adaptation problem by first generating potential movements and later selecting the most optimal according to the current situations and robot expertise and limitations.

In fact, this area of developing intelligent embodied systems that understand their own limitations has already been explored in the past. Prior works have considered physical self-awareness by predicting whether a robot will fall in the near future \cite{fallprediction:modelbased1, fallprediction:modelbased2, fallprediction:learning1, fallprediction:learning2, fallprediction:bilstm, fallprediction:1dcnn}. The main goal of these works is to avoid falls by adjusting its motion before high disturbances are encountered. Both model-based \cite{fallprediction:modelbased1, fallprediction:modelbased2} and learning-based approaches \cite{fallprediction:learning1, fallprediction:learning2, fallprediction:bilstm, fallprediction:1dcnn} have been considered for the task of fall prediction. For instance, \cite{fallprediction:bilstm, fallprediction:1dcnn} used deep-learning strategies to anticipate falls from the last observed robot states. However, all learning-based methods generated the training data using model-based controllers that were limited to simple movements, such as standing \cite{fallprediction:1dcnn} or walking \cite{fallprediction:bilstm}. They collect the falling examples by applying external force perturbations to the robot's torso in a simulator.  Consequently, their fall prediction approach was limited to very specific tasks (i.e., walking) and did not generalize to diverse human behaviors. On the contrary, this work proposes a system that understands the robot's capabilities when imitating any human reference provided, even high-dynamic movements such as backflips or handstands. We envision a general approach that allows a robot to imitate any human and any behavior commanded if it knows how or adapt the movement to its own expertise. 

Moreover, instead of just predicting a robot's potential fall, we anticipate how good a robot's imitation will be given a human reference. For that, we introduce a Self-AWare model, shorted as SAW, that infers a score to inform a robot on how smooth, safe, and faithful a robot's behavior will be if it decides to follow a human reference motion. Additionally, by integrating SAW with a real-time text-to-motion generator, such as MotionLCM \cite{dai2024motionlcm}, we can adapt the human's reference so that the robot imitates the high-level task commanded but improving the quality of the generated behavior. SAW can be used to generate and rank diverse potential robot behaviors and select the most adequate in advance, ensuring an optimal imitation. In conclusion, our efforts resulted in the following contributions:
\begin{enumerate}
    \item A transformer-based model to anticipate the robot's performance when following any human motion given as reference.
    \item A pipeline to optimize the robot's behavior, prevent falls, and enable non-expert users to control a bipedal robot using text or trajectory commands.
\end{enumerate}\textbf{}

\section{Related Works}
This section reviews the existing literature on robots imitating diverse human behavior, focusing on recent advancements and identifying current limitations. This analysis underscores the importance of integrating physical self-awareness in robots to facilitate their effective deployment in real world-applications.

\subsection{Imitation Learning}
As robots are increasingly utilized in more complex and unstructured environments, manually preprogramming their behavior or defining it through reward functions is becoming exceedingly difficult \cite{il:programmingbehaviors1, il:programmingbehaviors2}. Instead, imitation learning (IL) provides an avenue for teaching robots a desired behavior by simply demonstrating it. During the learning process, robots are provided with a dataset of expert demonstrations, and the goal is either to replicate them by mapping the observed states to actions, known as behavior cloning (BC) \cite{il:bc}, or to understand the underlying reward functions behind the expert's behavior with inverse RL (IRL) \cite{il:irl}. While IRL provides a more robust solution, its learning process is computationally more expensive and has lately struggled to scale to larger environments and replicate expert behaviors compared to BC \cite{il:irl}. In fact, BC has the advantage of being more efficient, as merely follows a traditional supervised learning task, but might suffer from a covariate shift problem \cite{il:covariateshiftproblem}. This problem arises due to the bias induced during training where the state distribution is led by an expert, while in testing the state is induced by its action  \cite{il:covariateshiftproblem2}. Different attempts \cite{il:bc, il:bc2, il:bc:error} have focused on solving the covariate shift problem, either by incorporating the human expert in the loop \cite{il:bc} or by limiting the agent actions to be in the distribution covered by the expert demonstrations \cite{il:bc:error}. For instance,  \cite{il:bc:error} learned to detect states that could lead to failures, and encourage BC to shift to known states. In fact, our work shares some common beliefs with \cite{il:bc:error} where anticipating potentially unstable states is crucial for ensuring optimal behaviors during BC, but we differ in nature. While \cite{il:bc:error} constraints the agent policy to be in the distribution of the expert demonstrations, we adapt the policy to the expertise of the agent itself. Our work builds upon behavior cloning but relaxes the imitation process, where the agent should perform a task optimally in its own domain. One similar motivation is proposed in \cite{il:bc:domaindiscrepancies}, where the goal is to overcome discrepancies between agent embodiments without forcing a shared latent domain.

\subsection{Human Imitation in Bipedal Robots}
Imitating human motion is no trivial task, especially for bipedal robots. Traditional methods in robot motion planning often rely on pre-defined movement patterns and control algorithms \cite{lee, dcm} inspired by human movements. These methods, though effective in controlled environments, struggle to adapt to the unpredictability of real-world settings. The lack of flexibility in these approaches results in robots that can perform specific tasks but fail when faced with novel or dynamically changing situations.

Reinforcement Learning (RL) has emerged as a powerful technique to address these limitations of traditional methods, showing high performance in animated characters  \cite{deepmimic, amp,perceptual}, quadruped robots, \cite{agarwal2023legged, pmlr-v205-fu23a} and recently in bipedal humanoids \cite{rl_bipedal1, rl_bipedal2, zhang2024whole, cheng2024express, he2024learning, zhang2024wococo, 2024ictrl}. All these approaches tackled behavior cloning through goal-conditioned RL \cite{goalcond_rl}, where a robot is trained to replicate a given human reference while considering environmental and physical constraints. For instance,  \cite{rl_bipedal1, rl_bipedal2} focused on training an RL agent to follow either velocity commands or jumps, respectively, and performed zero-shot transfer to a real Cassie robot. To gain more fine-grained control over the robot's behavior, \cite{zhang2024whole, zhang2024wococo} designed extensive rewards to overfit a policy for specific tasks in a bipedal humanoid robot. However, the ultimate goal of robot imitation is to enable a humanoid to follow any human demonstration, independent of its complexity and diversity. To address this task,  \cite{cheng2024express, he2024learning} focused on only imitating the upper-body part to account for more expressive robot behaviors, but only controlled the robot legs to ensure motion feasibility. Instead, I-CTRL \cite{2024ictrl} focused on whole-body humanoid imitation and showcased high performance in four different bipedal robots, generalizing over 10.000 different motions. For that, I-CTRL constrained the exploration phase during learning so that the produced physics-based behavior only modified the retargeted human reference within a defined margin, ensuring a highly visual resemblance. 

However, all the aforementioned works focused on blindly following a reference human motion with high precision, which could lead to falls or non-smooth behaviors when the intended reference behavior exceeds the robot's capabilities.  In fact, previous works filter out complex human behaviors (i.e., remove high jumps, backflips, ...) \cite{he2024learning, zhang2024wococo, 2024ictrl}, so that their generated behavior underperforms for complex reference motions, leading to falls. To overcome this issue, we propose to anticipate the resulting robot behavior of a trained policy when given a human reference. If a robot is aware that a commanded motion is out of its expertise and might lead to falls or poor imitation, the robot can decide to relax the reference motion constraints and further explore new reference behaviors that might lead to better performance. We described this ability as physical self-awareness, which refers to the robot's understanding of its own limitations and capabilities and enables the robot to make informed decisions.

\subsection{Self-Awareness in Robotics}
Psychologists describe self-awareness as the ability to become the object of one's attention \cite{saw:theory}, which entails a person's knowledge of themselves. This capability significantly influences human motivations, decisions, and intentions \cite{saw:intention}, either at the cognitive or physical level. For instance, \cite{saw:intention} states that when a human performs a movement, their action awareness is unconsciously monitoring discrepancies between the planned movement and the current state. If this error becomes significant, this awareness alerts a higher-level cognitive system to correct and replan this movement. Our work is inspired by this physical self-awareness and attempts to design a similar system for robots. Our developed Self-AWare model (SAW) continuously monitors discrepancies between the planned movements, the human references, and the current robot states. If SAW anticipates a potential discrepancy, such as a fall or a wrong robot imitation, our system proposes to replan the movement by consciously reasoning about the robot's limitations and capabilities.

In fact, preventing falls in robots, which can be considered as a subarea among physical self-awareness, has already been explored in the past. Early efforts focused on avoiding falls when small disturbances occurred using stabilizing controllers \cite{fallprediction:stab_control1, fallprediction:stab_control2}. However, when dealing with large disturbances, a robot requires more time to adjust its motion to avoid falling. Therefore, \cite{fallprediction:modelbased1, fallprediction:modelbased2} proposed to anticipate and prevent falls using model-based approaches. \cite{fallprediction:modelbased1} used a stand space inverted pendulum model for fall prediction, while \cite{fallprediction:modelbased2} modified the zero moment point (ZMP) for a simplified multi-rigid body model to predict falls in humans. Despite some advancements, these methods often oversimplify fall prediction by making assumptions that limit the robot's behavior to specific motions.

To address these limitations and build models that can better adapt to uncertainties, learning-based approaches were proposed \cite{fallprediction:learning1, fallprediction:learning2, fallprediction:bilstm, fallprediction:1dcnn}. \cite{fallprediction:learning1, fallprediction:learning2} hand-crafted robot state features to predict falls. Later, \cite{fallprediction:bilstm} inferred the likelihood of falling using a bidirectional Long-Short Term Memory (BiLSTM) network, which processed the evolution of robot sensor measurements such as the center of mass (CoM), the center of pressure (CoP) and the linear and angular momentum and its derivative. More recently, 
\cite{fallprediction:1dcnn} adopted a 1D convolutional neural network (1D-CNN) with the same goal. However, these learning-based methods generated the training data using model-based controllers that were limited to simple movements, such as standing \cite{fallprediction:1dcnn} and walking \cite{fallprediction:bilstm}. Consequently, their approach did not generalize to more diverse human behaviors a robot might want to imitate. On the contrary, we use I-CTRL \cite{2024ictrl} to generalize our SAW model to more complex and diverse motions, such as dancing, walking, running, and jumping.
\begin{figure*}[]
    \centering
    \includegraphics[width=0.95\textwidth]{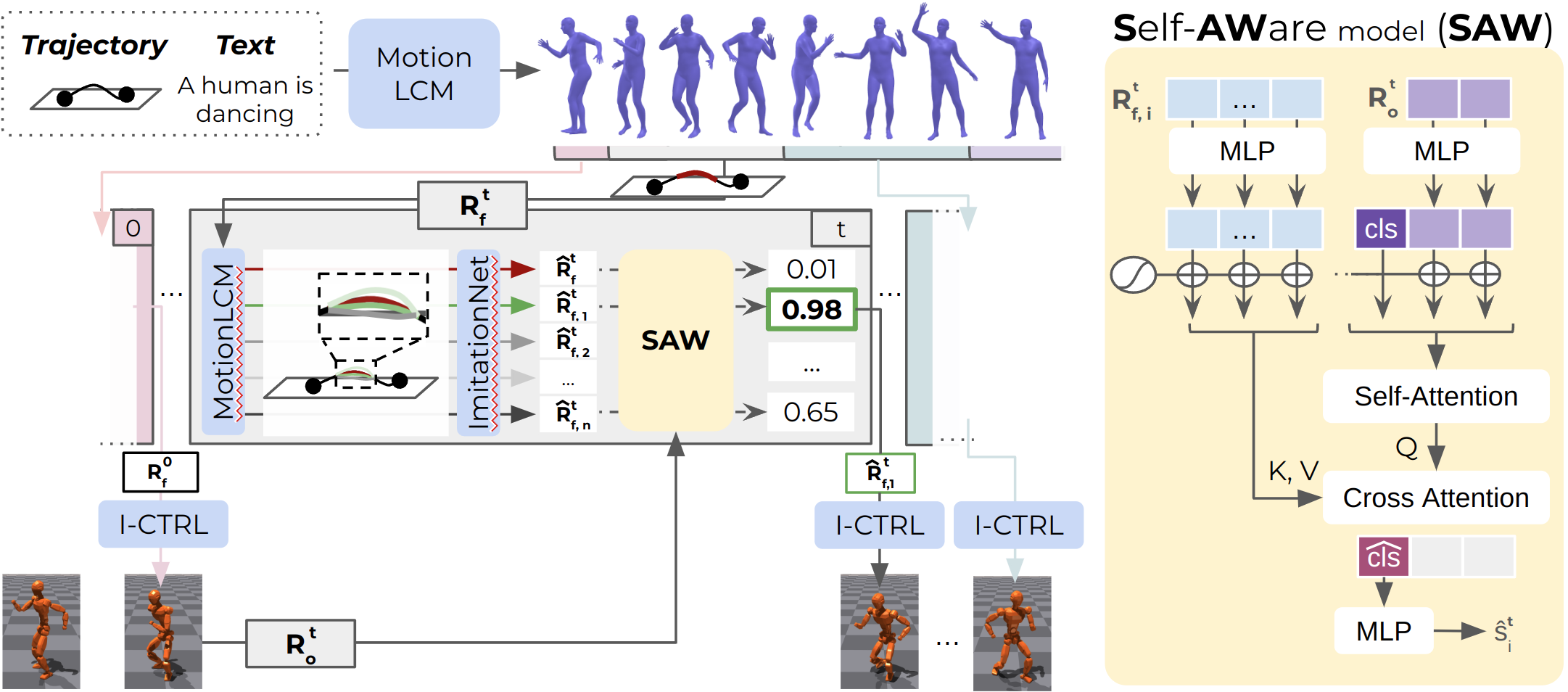}
    \caption{\textbf{Overview of our system for optimal robot behavior generation from high-level task commands with self-awareness.} Upon a high-level instruction, given as trajectory or natural language description, we generate a human reference $\mathbf{H_{r}}$ for the robot to follow using MotionLCM \cite{dai2024motionlcm}. Then, at each time horizon $t$, we assess the feasibility of the next future poses $\mathbf{R_{f}^t}$ to follow and optimize the robot behavior using the motion adapter module (shown in the gray center box). For that, MotionLCM is used first to generate potential new references with similar root trajectories and/or textual commands. Those references are retargeted to robot motions using ImitationNet, obtaining $[\mathbf{\hat{R}_{f, 1}^{t}}, \cdots, \mathbf{\hat{R}_{f, n}^{t}}]$. Our Self-AWare model (SAW, depicted in the light box on the right) ranks each edited references $\mathbf{\hat{R}_{f, i}^{t}}$ according to the robot's learned capabilities and the current robot states $\mathbf{\hat{R}_{o}^{t}}$. Thus, SAW infers a score vector $\hat{s}_i^t$ that describes how well the robot can follow a given reference $i$ at time $t$, and that can be summarized using a weighted sum to a single scalar per reference. The motion adapter then chooses the best reference and forwards it to I-CTRL \cite{2024ictrl}, which transfers this new reference to real robot commands $\mathbf{\hat{R}_{p}^{t}}$. }
    \label{fig:methodology}
\end{figure*}
Moreover, and contrary to prior works \cite{fallprediction:learning1, fallprediction:learning2, fallprediction:bilstm, fallprediction:1dcnn} that only focus on the observed robot states to predict falls, we incorporate the reference human motion to imitate as input to our model. For instance, when a bipedal robot imitates a human performing a high jump, the robot's self-awareness can recognize that the skill is out of its limits and adapt the behavior to perform a simpler jump that is feasible with the current abilities. Note that this prevention formulation differs from prior works on quadruped robots \cite{fallrecovery:quadruped:1} or small bipedal robots \cite{endtoend_rl_football} that train a specific policy for fall recovery, where the robot has already fallen. In our case, by accurately knowing the robot's limits and anticipating potential failures in advance, we can replan the robot's behavior to be optimal while remaining as close as possible to the intended movement.

\section{Methodology}

This section is structured as follows. First, we introduce the task of robot behavior generation and control from high-level commands with self-awareness. Then, we present our motion adaptation system with the Self-AWare model (SAW)  to tackle this task, which is illustrated in Fig. \ref{fig:methodology}. 

\subsection{Problem Formulation}
We envision the control of bipedal robots using high-level task commands ($c$), such as natural language descriptions or root trajectories. Instead of directly translating those commands to robot signals, we opted for using the human embodiment as a bridge and defining the task as an imitation problem. Therefore, our first subtask is to translate those high-level task commands $c$ to an appropriate human motion $\mathbf{H_{r}}$, which will serve as the reference for the robot to imitate. For that, we adopt a pre-trained MotionLCM \cite{dai2024motionlcm} as a text-to-motion generation model $f$, where ($f: c \longmapsto \mathbf{H_{r}}$). Our second subtask is to translate this human reference  $\mathbf{H_{r}}$ to a physics-based robot movement $\mathbf{R_{p}}$, such as ($g: \mathbf{H_{r}}\longmapsto \mathbf{R_{p}}$). In fact, we attain the description of $g$ as presented in \cite{2024ictrl}, where first a human-to-robot retargeting module $g_{h2r}$ translates a human reference $\mathbf{H_{r}}$ to a robot reference $\mathbf{R_{r}}$, only considering visual resemblance between the embodiments, and then a pre-trained RL module $g_{r2p}$ refines $\mathbf{R_{r}}$ to ensure plausibility under the real-world physics, generating $\mathbf{H_{p}}$. Following \cite{2024ictrl}, we define $g_{h2r}$ as ImitationNet  \cite{imitationnet} and $g_{r2p}$ as I-CTRL \cite{2024ictrl}.

However, due to the differences between kinematics and dynamics between humans and robots, as well as the error accumulation from ensembling multiple modules ($f$, $g_{h2r}$ and $g_{r2p}$), the resulting $\mathbf{R_{p}}$ might largely deviate from $\mathbf{H_{r}}$. Thus, we design a Self-AWare model (SAW) that learns how $g$ performs under different human references $\mathbf{H_{r}}$, and slightly adapts those references motions $\mathbf{\hat{H}_{r}}$ to achieve an optimal robot behavior that still satisfies the task command $c$.

Note that, given the use of multiple pre-trained models ($f$ is MotionLCM, $g_{h2r}$ is ImitationNet and $g_{r2p}$ is I-CTRL), $\mathbf{H_{r}}$, $\mathbf{R_{r}}$, and $\mathbf{R_{p}}$ are represented differently. A human motion  $\mathbf{H_{r}} \in  \mathbb{R}^{T \times J\times 3}$ is represented as a sequence of $T$ human poses, where each pose is defined as $J$ joints in Cartesian representation. On the contrary, $\mathbf{R_{r}} \in  \mathbb{R}^{T \times D_r}$   includes the root position $\mathbf{p}^{t} \in \mathbb{R}^{3}$ and orientation $\boldsymbol{\theta}^{t} \in \mathbb{R}^{4}$ in quaternion, as well as the robot joint angles $\mathbf{q}^{t} \in \mathbb{R}^{S}$, so that $D=3+4+S$. Similarly, $\mathbf{R_{r}} \in  \mathbb{R}^{T \times D_r+S}$ also includes the robot joint velocities  $\mathbf{\dot{q}}^{t} \in \mathbb{R}^{S}$.

\subsection{Motion Adaptation}\label{sec:methodology:madaptation}
We consider the motion adaptation task as finding a new human reference $\mathbf{\hat{H}_{r}}$ that ensures optimal robot behavior $\mathbf{R_{p}}$ while still resembling the original reference $\mathbf{H_{r}}$ and the task command $c$. Our reference adaptation block utilizes the last $T_o$ observed robot states $\mathbf{R_{p}^{t-T_o: t}} = [\mathbf{r}_p^{t-T_o}, \cdots, \mathbf{r}_p^{t}]$ to ensure that the new references are feasible with the current robot state, simplified as $\mathbf{R_{o}^{t}}$, and the subsequent $T_f$ human reference poses $\mathbf{H_{r}^{t: t+T_f}} = [\mathbf{h}_{r}^{t}, \cdots, \mathbf{h}_{r}^{t + T_f}]$ which we aim to adapt, simplified as $\mathbf{H_{f}^{t}}$, where $t$ represents the current time.

\begin{figure}[]
    \centering
    \includegraphics[width=0.47\textwidth]{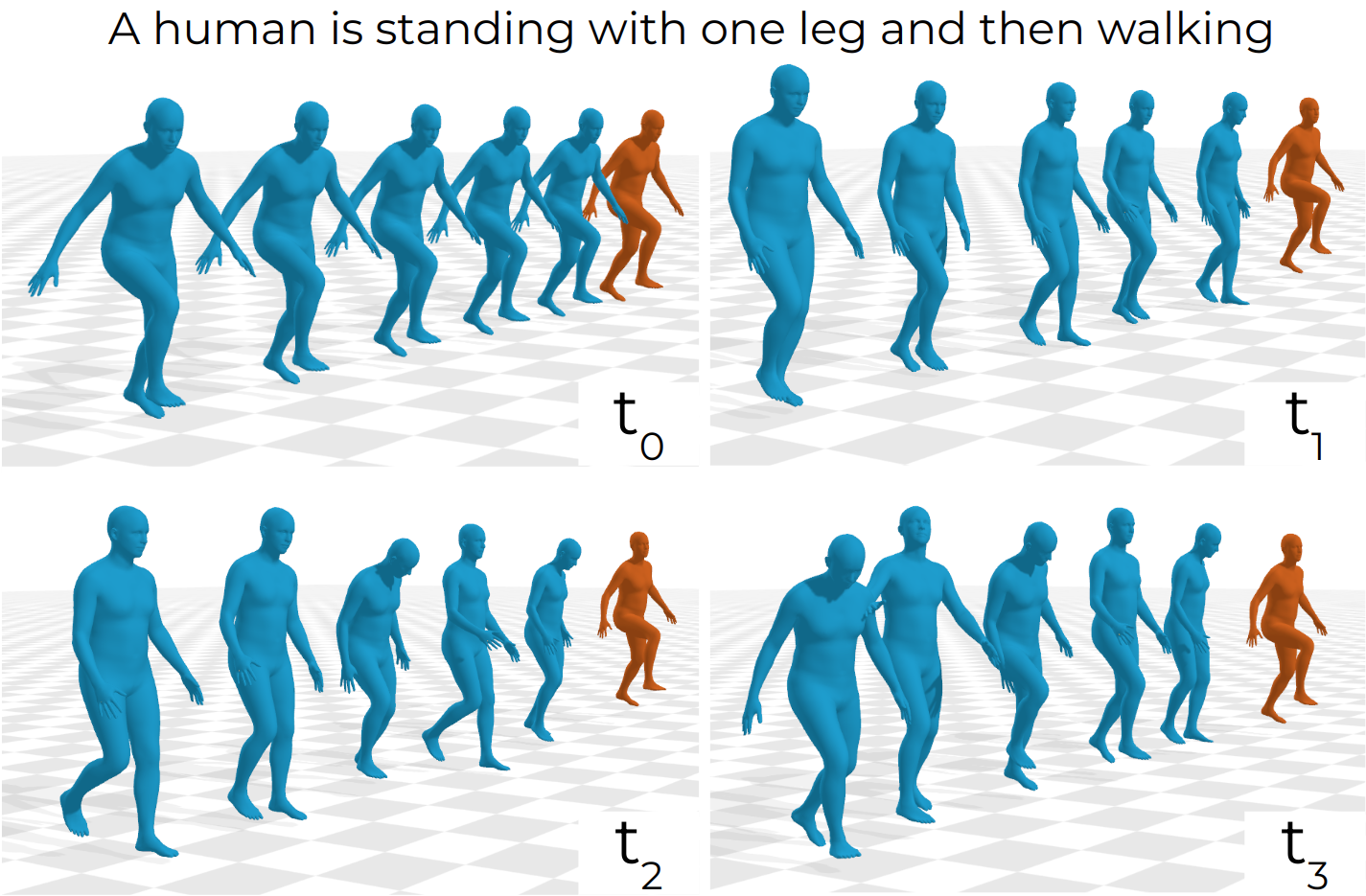}
    \caption{\textbf{Example of our Top-5 ranked edited motions given commanded human reference using our motion adaptation}. Here, the motion in orange is the original reference $\mathbf{H_{r}}$, while in blue, we show Top-5 edited references $[\mathbf{\hat{H}_{f, 1}^{t}}, \cdots, \mathbf{\hat{H}_{f, 5}^{t}}]$ at different times. Note that, as SAW predicts a high likelihood of falling for the `standing with one leg' for a long time, it tends to place the leg down and start `walking', as instructed by the high-level task command.}
    \label{fig:motionadapted}
\end{figure}

We define the motion adaptation similarly to a brainstorming process, where we first generate a set of $n$ modified reference motions $[\mathbf{\hat{H}_{f, 1}^{t}}, \cdots, \mathbf{\hat{H}_{f, n}^{t}}]$, and then we rank those potential motions to select the most optimal $\mathbf{\hat{H}_{f, i}}$ according to the robot feasibility. For that, we employ MotionLCM \cite{dai2024motionlcm} which allows joint and trajectory-level editing and can ensure that the edited reference motions start from the current pose $\mathbf{h}_{r}^{t}$ and closely approximate to $\mathbf{h}_{r}^{t + T_f}$ and its root trajectory. An example of this behavior is shown in Fig. \ref{fig:motionadapted}.

Later, we convert those edited references to robot references $[\mathbf{\hat{R}_{f, 1}^{t}}, \cdots, \mathbf{\hat{R}_{f, n}^{t}}]$ using ImitationNet \cite{imitationnet}.  Finally, each edited reference $\mathbf{\hat{R}_{f, i}^{t}}$ generated is then processed by SAW alongside the observed robot states $\mathbf{R_{o}^{t}}$, inferring a score $\hat{s}_i^t$ that informs on how optimal a reference $\mathbf{\hat{R}_{f, i}^{t}}$ is. By calculating a score $\hat{s}_i^t$ for each edited reference $\mathbf{R_{o}^{t}}$, we can rank and select the most suitable option that prioritizes avoiding falls and then optimize the quality of robot behavior. This chosen reference is followed until the SAW module identifies a better alternative for I-CTRL to guide the robot's actions. An overview of this process is shown in Fig. \ref{fig:methodology}.

\subsection{Self-AWare model (SAW)}\label{sec:methodology:saw}
This section provides a thorough description of our deep-learning-based Self-AWare model (SAW) for evaluating reference motions for optimal robot control. SAW is designed based on two main ideas: a robot's behavior might lead to failure as the current robot state is already not adequate or if the reference is unfeasible for the current situation. Therefore, we first encode the reference motion $\mathbf{\hat{R}_{f, i}^{t}}$ and the observed robot states $\mathbf{R_{o}^{t}}$ using individual multi-layer perceptrons (MLPs), which leads to $\mathbf{E_{f, i}^{t}}$ and $\mathbf{E_{o}^{t}}$ respectively. Next, we make use of the attention mechanisms \cite{transformer} to summarize both sequences. For that, we append a learnable class token $cls$ \cite{vit} to the $\mathbf{E_{o}^{t}}$ to aggregate all motion information during the transformer process. We add a sinusoidal positional embedding to $\mathbf{E_{o}^{t}}$  and forward it to a self-attention transformer model to learn the temporal relationships of the observed robot states, embedded as $\mathbf{\hat{E}_{o}^{t}}$. Finally, we condition this representation with the planned reference motion  $\mathbf{E_{f, i}^{t}}$ using cross-attention. Following \cite{vit}, we extract the appended $\widehat{cls}$ token from the cross-attention output which represents the expected robot behavior in the future if imitating the provided reference motion. $\widehat{cls}$ is then projected to a score vector $\hat{s}_i^{t}$ that assesses factors such as the likelihood of falling ($\widehat{fall}$), the smoothness of the generated robot's behavior ($\hat{A}_{\Ddot{q}}$), and the alignment of joint angles ($\hat{A}_q$), joint velocities ($\hat{A}_{\dot{q}}$), and root position ($\hat{A}_p$) and orientation ($\hat{A}_{\theta}$) with the reference. We define alignment as the mean square error between the reference state and the generated robot state, and smoothness as low robot joint accelerations (no jittering). Thus, a predicted score $s=[\widehat{fall}, \hat{A}_q, \hat{A}_{\dot{q}}, \hat{A}_{\Ddot{q}}, \hat{A}_p, \hat{A}_{\theta}] \in \mathbb{R}^{6}$.

\section{Experiments}

\begin{table*}[]
\centering
\resizebox{0.95\textwidth}{!}{%
\begin{tabular}{lccccccc}
 & Future Horizon & Fall Accuracy. (\%) $\uparrow$ & Smoothness Error $(A_{\Ddot{q}})$ $\downarrow$ & $A_{q}$ Error  $\downarrow$ & $A_{\dot{q}}$  Error $\downarrow$ & $A_{p}$ Error  $\downarrow$ & $A_{\theta}$ Error  $\downarrow$ \\ \midrule
w/o $\mathbf{R}_f$ & 1 & 97.70 & 0.093 & 0.0252 & 7.232 & 0.1136 & 0.0477 \\
w/o $\mathbf{R}_o$ & 1 &99.16 & 0.080 & 0.0396 & 5.760 & 0.1008 & 0.0225 \\
w/o C.A.           & 1 &99.14 & 0.074 & 0.0162 & 5.320 & 0.0944 & 0.0231 \\
SAW*               & 1 & 99.28 & 0.076 & 0.0156 & 5.248 & 0.0992 & 0.0207 \\
SAW                & 1 & \textbf{99.29} & \textbf{0.070} & \textbf{0.0144} &\textbf{4.592} & \textbf{0.0928} & \textbf{0.0066} \\ \midrule
SAW                & 2 & 99.17 & 0.057 & 0.0144 & 4.016 & 0.0976 & 0.0228\\
SAW                & 3 & 99.20 & 0.043 & 0.0126 & 4.000 & 0.1024 & 0.0261 \\
\end{tabular}
}
\caption{Quantitative evaluation of SAW for score prediction. The first section showcases the benefits of SAW over other variants in the architecture. The second section showcases the robustness of SAW over longer future horizons.}
\label{tab:ablationstudy}
\end{table*}

\subsection{Dataset generation}
To train the SAW model effectively, we automatically generate a dataset that accounts for the robot's ability to accurately mimic real-world motions. This involved synthesizing 100,000 human motion references using MotionLCM, incorporating 10,000 diverse textual annotations, with motion durations ranging from 3 to 12 seconds. For each reference, we simulated three unique JVRC-1 robot behaviors \cite{jvrc_robot} using I-CTRL within the IsaacGym simulator, culminating in a comprehensive dataset of 300,000 robot behaviors. To ensure a fair evaluation, in our validation and testing dataset, the robot falls in half of the motions.
Our SAW module observes 0.5 seconds in the past to score the reference of the next 1, 2, or 3 seconds. 

\subsection{Metrics}
To evaluate our SAW module we make use of simple mean-square error metrics between the predicted and ground-truth alignment scores ($\hat{A}_q, \hat{A}_{\dot{q}}, \hat{A}_p, \hat{A}_{\theta}$). We also use accuracy as a metric to assess the falling prediction ability of SAW. Note that a robot is considered to have fallen when the root height of the robot is lower than a predefined threshold, following the definition from \cite{2024ictrl}. 

\subsection{Robot}
In this work, we demonstrate the benefit of self-awareness on the JVRC-1 robot \cite{jvrc_robot}. This robot has 23 DoFs with a height of 140 centimeters and a weight of 62.2 kilograms.

\subsection{Quantitative and Qualitative Evaluation}

Due to the absence of existing benchmarks for score prediction, we conducted an ablation study to assess (a) different variations of our proposed architecture, and (b) different future horizons in which we anticipate the robot's behavior. The results in Table \ref{tab:ablationstudy} showcase the high accuracy of SAW in predicting if a robot will fall ($\approx 99\%$), as well as the high precision in determining the quality of the generated motion across multiple horizons.

First of all, we observe that when only considering the observed robot states (w/o $\mathbf{R}_f$), SAW underperforms when predicting the alignment of the robot with the reference, as expected. Likewise, when SAW only considers the reference motion (w/o $\mathbf{R}_o$), it can not ground the expected behavior to its current state, thus also failing to correctly predict an alignment. Later on, we evaluate two modifications on how to integrate $\mathbf{R}_f$ and $\mathbf{R}_o$ for the score prediction. First, we consider one independent self-attention block per sequence and compute the score from the concatenation of the $\widehat{cls}$ token from both sequences (w/o C.A.). Secondly, we inverted the sequences in the computation, such as the self-attention is computed over the reference and later conditioned on $\mathbf{E}_o$. Both variants achieve higher alignment errors and lower fall prediction accuracy. In general, we observe that the observed motion $\mathbf{R}_o$ has a higher influence on the future robot behavior quality rather than $\mathbf{R}_f$, as also shown in SAW w/o $\mathbf{R}_f$.

Additionally, we evaluated the performance of SAW  over longer future horizons (from 1 second to 3 seconds) as reported in Table \ref{tab:ablationstudy}, which showcases that we can still predict falls with high accuracy with low error in the robot performance behavior. We observed that the anticipation of the smoothness and joint angle and velocity alignment is independent of the future horizon, but rather depends strongly on the reference motion to follow. On the contrary, it is harder to anticipate the root alignment the longer the future horizon. 

\begin{figure*}[]
    \centering
    \includegraphics[width=0.97\textwidth]{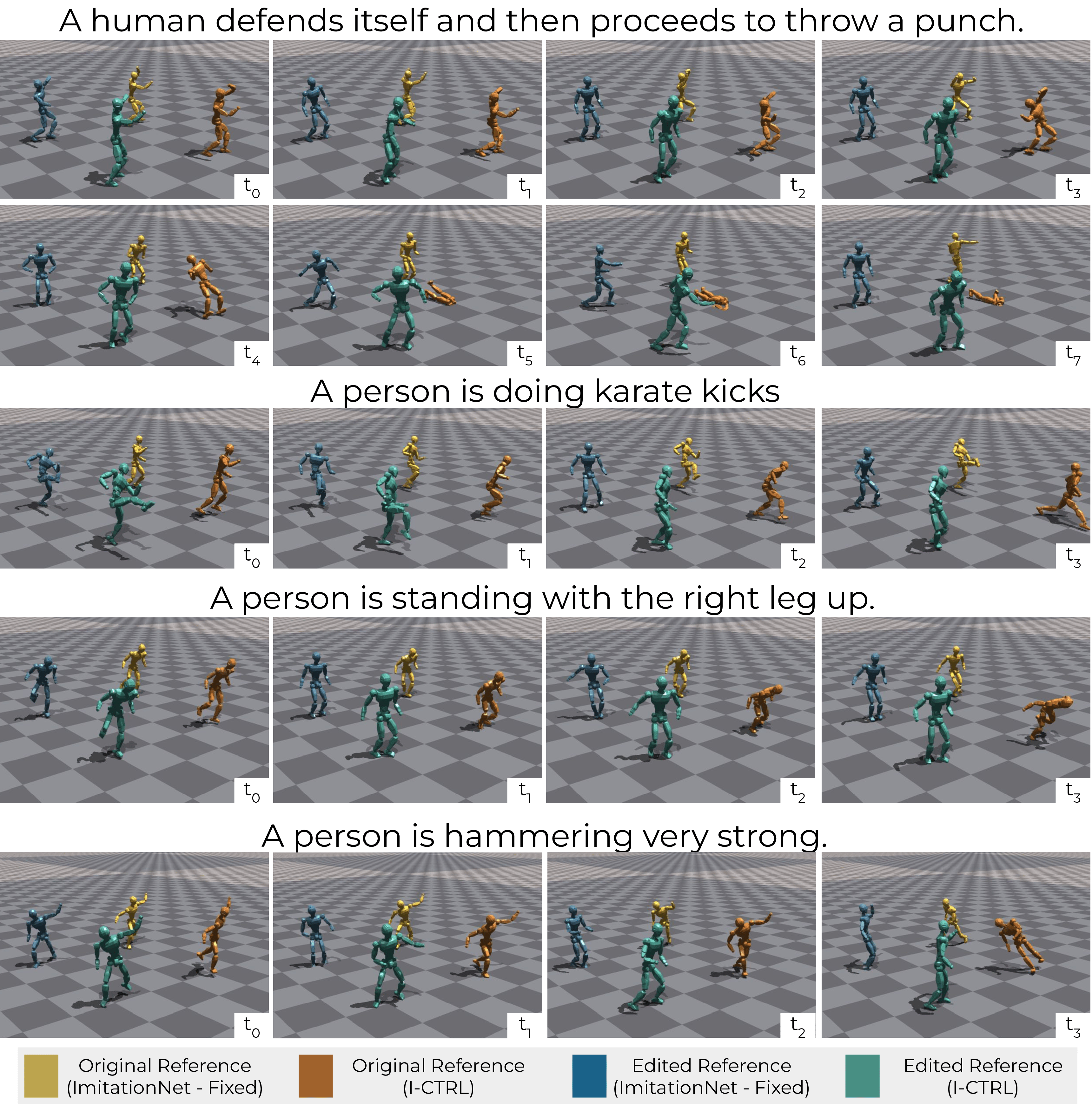}
    \caption{\textbf{Robot behavior comparison between using self-awareness (in dark and light blue) or not (in yellow and brown)}. Given a textual command, we use MotionLCM \cite{dai2024motionlcm} to generate a human reference, which we retarget to a JVRC-1 kinematics using ImitationNet \cite{imitationnet} to obtain a reference robot motion. Then, I-CTRL \cite{2024ictrl} is used to refine this robot reference (in yellow) to conform with real-world physics (in orange). However, if our SAW model anticipates poor imitation (i.e., falls), we adapt the reference motion to prevent wrong behaviors. For that, we use MotionLCM+ImitationNet to generate various new robot reference motions (in dark blue) and SAW selects the more appropriate, which we then command to I-CTRL to optimize the robot behavior (in light blue) to follow the initial commands (described here by text). Note that $t_k$ indicates different time instants of the motion sequences.}
    \label{fig:qualitativeresults}
\end{figure*}

Finally, we evaluated the overall system when the original robot's behavior (without self-awareness) was falling. Fig. \ref{fig:qualitativeresults} showcases the benefit of SAW when commanding motions out of the robot expertise. In general, our results demonstrate that our motion adaptation system can prevent 62\% falls while still following the same root trajectory.

\subsection{Limitations and future work}
Despite our SAW module's high performance in predicting the quality of the robot's motion when following a given reference ($\approx 99\%$), the end-to-end system only prevents 62\% of falls during tasks. We recognize that simply stopping the robot's motion and avoiding complex motions when a fall is anticipated could address these issues. However, this approach would prevent the robot from following the original commands of the human operator which was part of the scope of the motion adapter. Our analysis indicates that the primary cause of the system's limited fall prevention is the noisy reference motions generated by MotionLCM when conditioned on prior poses. These noisy references can lead to abrupt changes that I-CTRL cannot effectively predict. Additionally, we currently generate only 15 reference motions and rank the robot’s choices among them. Increasing this number to 30 or 50 would provide greater diversity, giving SAW better options to choose from, thereby potentially improving fall prevention.

\section{Conclusion}
Controlling any robot with high-level task commands often requires operators who understand the capabilities of the robots and limit the command accordingly. Instead, we propose to provide robots with self-awareness so that they can autonomously adjust the commanded instruction to their own ability. Our end-to-end system converts natural language and trajectory instructions into a human motion reference for the robot to follow. To mitigate the risk of poor robot performance due to strict adherence to commands, we introduce the Self-AWare model (SAW). Our SAW module is able to anticipate a fall with $\approx 99\%$ accuracy at different future horizons, and rank different references with high precision. Finally, we extend our self-awareness model with a motion adapter system that enables the robot to intelligently select the optimal reference motion according to the current robot states and capabilities, enhancing performance and preventing falls.

\newpage
\section*{ACKNOWLEDGMENT}
This work is funded by Marie Sklodowska-Curie Action Horizon 2020 (Grant agreement No. 955778) for the EU projects of PERSEO and INVERSE (GA nr. 101136067). 


\bibliographystyle{unsrt}
\bibliography{relatedworks}

\end{document}